8-week report for the Academies' Summer Research Fellowship Programme 2019

# Automatic Speech Recognition for Hindi

Anish Saha

Vellore Institute of Technology, Vellore 632014

Prof. A.G. Ramakrishnan

Indian Institute of Science, Bengaluru 560012

## Abstract

Automatic speech recognition is a leading problem of computational linguistics that involves development of technologies that enable the recognition and transcription of spoken language into text by computers. It incorporates knowledge and research in the fields of linguistics and machine learning. All such models which capture the correlation between the speech audio and the transcripts or vice versa, commonly known as supervised learning, require handling of real and unrestricted text, directly or indirectly. For example, text-to-speech systems directly need to work on real text, whereas automatic speech recognition systems depend on language models that are trained on a huge corpus of text. To properly train a predictive model, the transcribed data must meet exceptionally high-quality standards. Hence, the first part of the research involved designing a web application, using JavaScript and Node.js, which effectively handles large amounts of audio files and their transcriptions, and performs collaborative human correction of the ASR transcripts. The web app is designed to operate in real-time, using a client-server architecture. The second leg of research involved designing a web interface for speech recognition. This involved specific recording of 16 kHz mono channel audio from any device hosting the web app, and performs voice activity detection (VAD), and then sends the audio to the recognition engine. VAD is a technique used in speech processing in which the presence or absence of human speech is detected which can facilitate efficient speech processing or deactivate some processes during non-speech intervals of an audio session to avoid unnecessary coding & transmission of silence packets in VoIP applications, saving computation and network bandwidth. The final work involved testing a neural network which functions to align the speech signal to HMM states accurately. This was followed by testing a novel implementation of backpropagation utilizing prior statistics of node co-activations.

**Keywords or phrases**: Speech transcription, web application, language modelling, voice activity detection, hidden Markov model, neural networks, co-activation modeling

# Abbreviations

| HMM | Hidden Markov Model |
| --- | --- |
| ASR | Automatic Speech Recognition |
| VAD | Voice Activity Detection |
| G2P | Grapheme-to-Phoneme |
| TTS | Text-to-Speech |
| DNN | Deep Neural Network |

# 1 INTRODUCTION
## 1.1 Background

Hindi, being the official language of India, is the most commonly spoken language. It is also the fifth most spoken language in the world, with 33 percent of Indian population speaking it as their first language and as a second language by 120 million more.

The prevalance of Hindi coupled with a small percent of Indians using English for communication and literacy makes the availability of a commercial grade automatic speech recognition system for the Hindi language necessary.

## 1.2 Problem Statement

Good quality ASR systems for Hindi, or any Indian language for that matter, that can be used for real time deployment are not available. Though a number of prototypes are created and research is being carried out, for Indian language systems such as [1] [2], none of these are of quality that can be compared to commercial grade ASR systems for languages like English, German and French. The primary reason for this is that developing any system for a new language needs huge amount of transcribed speech corpus for that language. Furthermore, we need close collaboration between linguists and computer scientists, for resolving language specific issues.

Such a large corpus must be collected from multiple speakers of varying origins and dialects to adequately capture the diversity in a language and train the system well, allowing it to cope with any speaker, adapting to his style. Other language specific issues involve problems such as grapheme-to-phoneme (G2P) conversion, schwa deletion and compound words in Hindi [3].

## 1.3 Objectives of the research

1. Designing a web application, using JavaScript and Node.js, which effectively handles large amounts of audio files and their transcriptions, and performs collaborative human correction of the ASR transcripts. The web app is designed to operate in real-time, using a client-server architecture, whereby information about each manual edit is posted to the server and a change record is maintained.
2. Designing a web interface for speech recognition. This involved specific recording of 16 kHz mono channel audio from any device hosting the web app, and performs voice activity detection to detect presence or absence of human speech and deactivate some processes during non-speech intervals.

3. Collection of speech data and transcripts from 35 Hindi speakers to adapt and train the ASR model.
4. Testing a neural network which functions to align with and map the speech signal to HMM states accurately.
5. Developing a G2P conversion system and lexicon model to generate phonemes from Hindi transcripts, adhering to language specific rules.
6. Testing a novel implementation of backpropagation utilizing prior statistics of node co-activations.

# 2 LITERATURE REVIEW

In speech files, about half of the conversation contains speech, and the remaining part contains pauses or silence intervals. The inclusion of non-speech intervals in the speech files necessitates voice activity detection because these intervals do not contain any speaker information. Speech/non-speech detection can be formulated as a statistical hypothesis problem aimed at determining to which class a given speech segment belongs. However, a high level of background noise can cause numerous detection errors, because the noise partly or completely masks the speech signal [4]. Traditionally, VAD uses periodicity measure, zero-crossing rate, pitch, energy, spectrum analysis [5] [6], higher order statistics in the LPC residual domain [7]. More sophisticated VAD techniques have been proposed for real-time speech transmission over the internet.

All areas of language and speech technology, directly or indirectly, requires handling of real (unrestricted) text. Novel approaches to text normalization, wherein tokenization and initial token classification are combined into one stage, using a lexical analyser, followed by a second level of token sense disambiguation using decision lists and decision trees have been proposed [8].

Hindi ASR systems require good quality phonetizers for the grapheme-to-phoneme conversion. Various language related problems such as schwa-deletion rules and compound word detection are required to be handled by the lexicon model. [9]

presents appropriate solutions which overcome these issues by implementing a Hindi TTS system using the Festival framework and a generic G2P converter.

In order to overcome issues that plagues practical applications, especially ASR, utilizing deep neural network (DNN) systems is the mismatch between the training and real-world testing conditions, since the testing acoustic environment is difficult to pre-estimate, the performance generally suffers in the presence of unexpected noise. A proposed method is to adapt the model layer by layer to minimize the distance between the co-activation statistics of nodes in matched versus mismatched conditions (at speaker level) using knowledge of the statistics of its output and hidden layer activations [10].

# 3 METHODOLOGY

All speech processing applications, such as ASR and TTS, directly or indirectly, require handling of real (unrestricted) text. In order to manage the huge amounts of speech and text (transcript) data, a web application was designed, using JavaScript and Node.js, which effectively handles large amounts of audio files and their transcriptions, and performs collaborative human correction of the ASR transcripts. The web app was designed to operate in real-time, using a client-server architecture and a database hosted on MongoDB. The web app implemented a REST API which handled parallel logins from multiple users, each of whom had been allocated a user ID, passwords and language ID. The logged in user needs to load all wav audio files from the local machine into the web app which would then present the user with the transcripts of the corresponding audio files. Any corrections made to the transcripts were saved as txt files in the server. The app also allowed users to automatically convert numbers to words and expand abbreviations by just selecting the text and requesting the server for the conversion. Information about each manual edit was posted to the server and a change record was maintained, which could be used to provide incentives to the volunteers based on the updates made in the editing pipeline.

A web interface was created for speech recognition, which provided an user interface allowing users to record their voice and request the server to recognize and transcribe it to text. This involved specific recording of 16 kHz mono channel audio from any device hosting the web app. This was achieved by down-sampling the recorded data by omitting excess data points from the recorded speech vector. Speech files contains pauses or silence intervals. This inclusion of non-speech intervals in the speech files necessitates voice activity detection because these intervals do not contain any speaker information. The web interface was thus designed to performs voice activity detection, also known as speech activity detection to detect the presence or absence of human speech and can facilitate efficient speech processing by deactivating some processes during non-speech intervals of an audio session to avoid unnecessary coding & transmission of silence packets in VoIP applications, saving on computation and on network bandwidth. This was achieved by using the time-domain data from the speech vector in a fixed FFT window size, and checking the highest amplitude reached within that window. If it exceeds a threshold, determined experimentally, the vector contained speech otherwise it consisted of silence which could be ignored.

ASR requires a considerable amount of speech data to train its model. This required collection of Hindi speech samples from multiple speakers of different origins and dialects. The volunteers were required to speak out Hindi sentences into a microphone which recorded the audio in 16 kHz mono channel wav format. The volunteers were given transcripts containing 120 sentences written in Hindi out of which 6 meaningless sentences were common for all speakers (used for speaker adaptation by the system) and 114 meaningful sentences different for each speaker. The audio was recorded in a noise free environment with no stammering or fumbling made by the speakers, taking 45 minutes on an average per speaker. All speakers were above the age of 16 and mostly comprised of fluent Hindi speakers, with considerable data collected from Hindi professors and Hindi major students. This resulted in almost clean and good quality data which could be used by the ASR system for training and testing.

The utterances files consist of sentences, which are made of words - consisting of multiple phones. Each phone corresponds to a sequence of HMM states, with words and sentences being formed by emission of feature vectors a chain of certain HMM states. A single phone consists of a rising state, a stable state and a falling state per HMM. Therefore, each utterance needs to be segmented and mapped to appropriate alignments (i.e., HMM states). The final stage of this processing was implemented by using a DNN. The DNN had 7 layers consisting of both trainable and non-trainable layers and sigmoid activations. This architecture was tested thoroughly by cross checking each hidden layer's output and activation and the gradients being calculated at each layer during backpropagation (for gradient descent based optimization). This was followed by implementing and testing a novel approach for network adaptation where the neural network uses prior knowledge of the statistics of its output and hidden layer activations to update its parameters online. Since individual nodes in a neural network become selective to particular phonetic features, the co-activation of the hidden layers should encode information about the distributions of phones present in a speech signal. The co-activation consisted of the mean of the hidden layer activation outputs and their inverse covariance matrix. These were incorporated into the gradient descent formulation as a separate term in the loss function so as to improve the accuracy and convergence rates, by minimizing the distance between the co-activation statistics of nodes in matched versus mismatched conditions.

Finally, Hindi ASR systems require good quality phonetizers for the grapheme-to-phoneme conversion. Various language related problems such as schwa-deletion rules and compound word detection required handling. A lexicon model was built in Java using the Devanagari unicodes and considering the schwa-deletion rules presented in the paper [9].

# 4 RESULTS AND DISCUSSION

A web application was designed, using JavaScript and Node.js, which effectively handled large amounts of audio files and their transcriptions and allowed collaborative human correction of the ASR transcripts by volunteers who were assigned audio files. The web app maintained information about each manual edit in a MongoDB database coupled with scripts that were used to maintain and backup the batabase.

For the existing automatic speech recognition system for Tamil, a web user interface was created. This allowed users to go to the website and record speech as 16 kHz mono channel audio from any device, with any duration of silence in between speech. The voice activity detection feature to detect presence or absence of human speech and deactivate some processes during non-speech intervals allowed the user to experience continous speech recognition once a button was pressed.

Speech data and transcripts for the Hindi language were collected from speakers to adapt, train and test the ASR model. 120 sentences of Hindi text were recorded each from 30 speakers. The collected data was free from background noise and consisted of mostly clean and accurate data.

The DNN which functions to align and map the speech utterance to HMM states was tested thoroughly, matching and verifying the outputs of each hidden layer along with the gradients obtained during backpropagation gradient descent.

A G2P conversion system and lexicon model was written in Java, to generate phonemes from Hindi transcripts, adhering to language specific rules such as schwa inclusion and deletion along with handling a multitude of vocalic and nasal variations of consonant sounds in Hindi.

A novel implementation of backpropagation utilizing prior statistics of node co-activations was completed, and the statistics such as mean and co-variance of hidden layer activations were manually calculated and matched. The technique, as expected, performed better than vanilla gradient descent with greater reduction in the error metric and improvement in classification accuracy per epoch of training.

# 5 CONCLUSION

Speech recognition technology which is an increasingly popular concept in recent years attracts attension from organizations to individuals; the technology being widely used for the various advantages it provides. It brings the ability for a machine to listen and understand what people are talking or what user are commanding. The research and implementation of an entire speech recognition model for the Hindi language provided great insight into the technology and future scope for research. Also, the web app was successfully used to correct transcripts by volunteers; and the web interface appropriately recorded and displayed speech transcripts that were recognized by the system.

# ACKNOWLEDGEMENTS

I would like to thank Dr. A G Ramakrishnan (Professor, EE department, IISc) and Mr. Madhavaraj A (Research Scholar, MILE Lab EE department, IISc) my mentor, for this opportunity to work as a summer research intern at MILE lab IISc Bangalore. Also I would like to thank my mentor for his guidance and inputs throughout the course of this work.